# Phase recovery and holographic image reconstruction using deep learning in neural networks

Yair Rivenson[1,2,3]†, Yibo Zhang[1,2,3]†, Harun Günaydın[1], Da Teng[1,4], Aydogan Ozcan[1,2,3,5]*

**Affiliations:**

[1]Electrical Engineering Department, University of California, Los Angeles, CA, 90095, USA.

[2]Bioengineering Department, University of California, Los Angeles, CA, 90095, USA.

[3]California NanoSystems Institute (CNSI), University of California, Los Angeles, CA, 90095, USA.

[4]Computer Science Department, University of California, Los Angeles, CA, 90095, USA

[5]Department of Surgery, David Geffen School of Medicine, University of California, Los Angeles, CA, 90095, USA.

*Correspondence: ozcan@ucla.edu

Address: 420 Westwood Plaza, Engineering IV Building, UCLA, Los Angeles, CA 90095, USA

Tel: +1(310)825-0915

Fax: +1(310)206-4685

†Equal contributing authors.

**Abstract**

Phase recovery from intensity-only measurements forms the heart of coherent imaging techniques and holography. Here we demonstrate that a neural network can learn to perform phase recovery and holographic image reconstruction after appropriate training. This deep learning-based approach provides an entirely new framework to conduct holographic imaging by rapidly eliminating twin-image and self-interference related spatial artifacts. Compared to existing approaches, this neural network based method is significantly faster to compute, and reconstructs improved phase and amplitude images of the objects using only one hologram, i.e., requires less number of measurements in addition to being computationally faster. We validated this method by reconstructing phase and amplitude images of various samples, including blood and Pap smears, and tissue sections. These results are broadly applicable to any phase recovery problem, and highlight that through machine learning challenging problems in imaging science can be overcome, providing new avenues to design powerful computational imaging systems.

**Introduction**

Opto-electronic sensor-arrays, e.g., charge-coupled devices (CCDs) or complementary metal-oxide-semiconductor (CMOS) based imagers, are only sensitive to the intensity of light, and therefore phase information of the objects or the diffracted light waves cannot be directly recorded using such imagers. Phase recovery from intensity-only measurements has emerged as an important field to recover this lost phase information at the detection process, enabling the reconstruction of the phase and amplitude images of specimen using various approaches[1-13]. In fact, Gabor's original in-line holography system[14], where the diffracted light from the object interferes with the background light that is directly transmitted, is an important example where phase recovery is needed to separate the twin-image and self-interference related spatial artifacts from the real image of the sample. In various implementations, to improve the performance of phase recovery and image reconstruction processes, additional intensity information is recorded, e.g., by scanning the illumination source aperture[15–18], sample-to-sensor distance[19–22], wavelength of illumination[23,24], or phase front of the reference beam[25–28], among other methods[29–34], all of which utilize additional physical constraints and intensity measurements to robustly retrieve the missing phase information based on an analytical and/or iterative solution that satisfies the wave equation. Some of these phase retrieval techniques have enabled new discoveries in different fields[35–38].

Here we report a convolutional neural network-based method, trained through deep learning[39,40], that can perform phase recovery and holographic image reconstruction using a single intensity-only hologram. Deep learning is a machine learning technique that uses a multi-layered artificial neural network for data modeling, analysis and decision making, and it has shown considerable success in areas where large amounts of data are available. Our deep learning-enabled coherent image reconstruction framework is very fast to compute, taking e.g., ~3.11 sec on a graphics processing unit (GPU) based laptop computer to recover phase and amplitude images of a specimen over a field-of-view of 1 mm$^2$, containing ~7.3 megapixels in each image channel (amplitude and phase). We validated this approach by reconstructing complex-valued images of various samples including e.g., blood and Papanicolaou (Pap) smears as well as thin sections of human tissue samples, all of which demonstrated successful elimination of the twin-image and self-interference related spatial artifacts that arise due to lost phase information at the hologram detection process. Stated differently, after its training, the convolutional neural network learned to extract and separate the spatial features of the real image from the features of the twin-image and other undesired interference terms for both the phase and amplitude channels of the object. Remarkably, this deep learning based phase recovery and holographic image reconstruction have been achieved without any modeling of light-matter interaction or wave interference. However, this does not imply that the presented approach entirely ignores the physics of light-matter interaction and holographic imaging, which is in fact statistically inferred through deep learning in a convolutional neural network by using a large number of microscopic images as gold standard in the training phase. This training and statistical optimization of the neural network only need to be performed once, and can be considered as part of a blind reconstruction framework that performs phase recovery and holographic image reconstruction using a single input, i.e., an intensity-only hologram of the object. This framework opens up a myriad of opportunities to design fundamentally new coherent imaging systems, and can be broadly applicable to any phase recovery problem, spanning different parts of the electromagnetic spectrum, including e.g., visible wavelengths as well as X-rays[26,28,41,42].

**Results**

Our deep neural network approach for phase retrieval and holographic image reconstruction is schematically described in Fig. 1 (also see Supplementary Figs. 1-3). In this work, we chose to demonstrate the proposed framework using lensfree digital in-line holography of transmissive samples including human tissue sections, blood and Pap smears (see the Methods Section). Due to the dense and connected nature of these samples that we imaged, their holographic in-line imaging requires the acquisition of multiple holograms for accurate and artifact-free object recovery[43]. A schematic of our experimental set-up is shown in Supplementary Fig. 4, where the sample is positioned very close to a CMOS sensor chip, with < 1 mm sample-to-sensor distance, which provides an important advantage in terms of the sample field of view that can be imaged. However, due to this relatively short sample-to-sensor distance the twin-image artifact of in-line holography, which is a result of the lost phase information, is rather strong and severely obstructs the spatial features of the sample in both the amplitude and phase channels, as also illustrated in Figs. 1-2.

The first step in our deep learning-based phase retrieval and holographic image reconstruction framework involves "training" of the neural network, i.e., learning the statistical transformation between a complex-valued image that results from the back-propagation of a *single* hologram intensity of the object and the same object's image that is reconstructed using a multi-height phase retrieval algorithm (treated as gold standard for the training phase) using 8 hologram intensities acquired at different sample-to-sensor distances (see the Methods section as well as Supplementary Information). As illustrated in Figs. 1-3, a simple back-propagation of the object's hologram, without phase retrieval, contains severe twin-image and self-interference related artifacts, hiding the phase and amplitude information of the object. This training/learning process (which needs to be performed only once) results in a ***fixed*** deep neural network that is used to ***blindly*** reconstruct, using a single hologram intensity, phase and amplitude images of any object, free from twin-image and other undesired interference related artifacts.

In our holographic imaging experiments, we used three different types of samples, i.e., blood smears, Pap smears and breast tissue sections, and separately trained three convolutional neural networks for each sample type, although the network architecture was identical in each case as shown in Fig. 1. To avoid over-fitting of the neural network, we stopped the training once the deep neural network performance on the validation image set (which is entirely different than the training image set as well as the blind testing image set) starts to reduce. We also accordingly made the network compact and applied pooling approaches[44]. Following this training process, each deep neural network was ***blindly*** tested with different objects that were ***not*** used in the training or validation image sets. Figs. 1, 2 and 3 demonstrate some of these neural network-based blind reconstruction results for Pap smears, breast tissue sections and blood smears. These reconstructed phase and amplitude images clearly demonstrate the success of our deep neural network-based holographic image reconstruction approach to blindly infer artifact-free phase and amplitude images of the objects, matching the performance of multi-height phase recovery. Table 1 further compares the structural similarity[45] (SSIM) of our neural network output images (using a single input hologram, i.e., $N_{holo} = 1$) against the results obtained with a traditional multi-height phase retrieval algorithm using multiple holograms (i.e., $N_{holo} = 2, 3,…,8$) acquired at different sample-to-sensor distances. A comparison of the SSIM index values reported in Table 1 suggests that the imaging performance of the deep neural network using a single hologram is comparable to that of multi-height phase retrieval, e.g., closely matching the SSIM performance of $N_{holo} = 2$ for both Pap smear and breast tissue samples, and the SSIM performance of $N_{holo} = 3$ for blood smear samples. In other words, the deep neural network-based reconstruction approach reduces the number of holograms that needs be acquired by 2-3 times. In addition to this reduction in the number of holograms, the computation time for holographic reconstruction using a neural network is also improved by more than 3- and 4-fold compared to multi-height phase retrieval with $N_{holo} = 2$ and $N_{holo} = 3$, respectively (see Table 2).

The phase retrieval performance of our neural network is further demonstrated by imaging red blood cells (RBCs) in a whole blood smear. Using the reconstructed phase images of RBCs, the relative phase delay with respect to the background (where no cells are present) is calculated to reveal the *phase integral* per RBC (given in units of rad·µm$^2$ - see Supplementary Information for details), which is directly proportional to the volume of each cell, *V*. In Fig. 3(a), we compare the phase integral values of 127 RBCs in a given region of interest, which were calculated using the phase images of the network input, network output, and the multi-height phase recovery image obtained with $N_{holo} = 8$. Due to the twin-image and other self-interference related spatial artifacts, the effective cell volume and the phase integral values calculated using the network input image demonstrated a highly random behavior, as shown with the scattered blue dots in Fig. 3(a), which is significantly improved by the network output, shown with the red dots in the same figure.

Next, to evaluate the tolerance of the deep neural network and its holographic reconstruction framework to axial defocusing, we digitally back-propagated the hologram intensity of a breast tissue section to different depths, i.e., defocusing distances within a range of $z = [-20\ \mu m, +20\ \mu m]$ with $\Delta z = 1\ \mu m$ increments. After this defocusing, we then fed each resulting complex-valued image as input to the same fixed neural network (which was trained by using in-focus images, i.e., $z = 0\ \mu m$). The amplitude SSIM index of each network output was evaluated with respect to the multi-height phase recovery image with $N_{holo} = 8$ used as the reference (see Fig. 4). Although the deep neural network was trained with in-focus images, Fig. 4 clearly demonstrates the ability of the network to blindly reconstruct defocused holographic images with a negligible drop in image quality across the imaging system's depth of field, which is ~4 µm.

**Discussion**

In a digital in-line hologram, the intensity of the light incident on the sensor array can be written as:

$$I(x,y) = |A + a(x,y)|^2 = |A|^2 + |a(x,y)|^2 + A^* a(x,y) + Aa^*(x,y) \tag{1}$$

where *A* is the uniform reference wave that is directly transmitted, and *a(x,y)* is the complex-valued light wave that is scattered by the sample. Under plane wave illumination, we can assume *A* to have zero phase at the detection plane, without loss of generality, i.e., $A = |A|$. For a weakly scattering object, one can potentially ignore the self-interference term, $|a(x,y)|^2$, compared to the other terms in equation (1) since $|a(x,y)| \ll A$. As detailed in our Supplementary Information, none of the samples that we imaged in this work satisfies this weakly scattering assumption, i.e., the root-mean-squared (RMS) modulus of the scattered wave was measured to be approximately 28%, 34% and 37% of the reference wave RMS modulus for breast tissue, Pap smear and blood smear samples, respectively. That is why, for in-line holographic imaging of such strongly-scattering and structurally-dense samples, self-interference related terms, in addition to twin-image, form strong image artifacts in both phase and amplitude channels of the sample, making it nearly impossible to apply object support-based constraints for phase retrieval. This necessitates additional holographic measurements for traditional phase recovery and holographic image reconstruction methods, such as the multi-height phase recovery approach that we used for comparison in this work. Without increasing the number of holographic measurements, our deep neural network-based phase retrieval technique can learn to separate/clean phase and amplitude images of the objects from twin-image and self-interference related spatial artifacts as illustrated in Figs. 1-3. In principle one could also use off-axis interferometry[46,47] for imaging of such strongly scattering samples. However, this would create a penalty for resolution or field-of-view of the reconstructed images due to the reduction in the space-bandwidth product of an off-axis imaging system.

Another important property of this deep neural network-based holographic reconstruction framework is the fact that it significantly suppresses out-of-focus interference artifacts, which frequently appear in holographic images due to e.g., dust particles or other imperfections in various surfaces or optical components of the imaging set-up – some of these naturally occurring artifacts are also highlighted in Fig. 2(f,g,n,o) with yellow arrows, which were cleaned in the corresponding network output images, Fig. 2(d,e,l,m). This property stems from the fact that such holographic out-of-focus interference artifacts, from the perspective of our trained neural network, fall into the same category as twin-image artifact due to the spatial defocusing operation, helping the trained network reject such artifacts in the reconstruction process. This is especially important for coherent imaging systems since various unwanted particles and features form holographic fringes on the sensor plane, superimposing on the object's hologram, which degrade the perceived image quality after image reconstruction.

In this work we used the exact same neural network architecture depicted in Fig. 1 and Supplementary Figs. 1-2 for all the object types, and based on this design we separately trained the convolutional neural network for different types of objects (e.g., breast tissue vs. Pap smear), which was then fixed after the training process to blindly reconstruct phase and amplitude images of any object of the same type. This does not pose a limitation since in most imaging experiments the type of the sample is known, although its microscopic features are unknown and need to be revealed by a microscope. This is certainly the case for biomedical imaging and pathology since the samples are prepared (e.g., stained and fixed) with the correct procedures, tailored for the type of the sample. Therefore, the use of an appropriately trained neural network for a given type of sample can be considered very well aligned with traditional uses of digital microscopy tools.

Having emphasized this point, we also created and tested a universal neural network that can reconstruct different types of objects after its training, still based on the same architecture used in our earlier networks. To handle different object types using a single neural network, we increased the number of feature maps in each convolutional layer from 16 to 32 (see the Supplementary Information), which also increased the complexity of the network, leading to increased training times, while the reconstruction runtime (after the network is fixed) marginally increased from e.g., 6.45 sec to 7.85 sec for a field-of-view of 1 mm$^2$ (see Table 2). Table 1 also compares the SSIM index values that are achieved using this universal network, which performed very similar to individual object type specific networks. A further comparison of holographic image reconstructions that are achieved by this universal network against object type specific networks is also provided in Figure 5, revealing the same conclusion as in Table 1.

## Methods

**Multi-height phase recovery**

To generate ground truth amplitude and phase images used to train the neural network, phase retrieval was achieved by using a multi-height phase recovery method[19,21,22]. For this purpose, the image sensor is shifted in the z direction away from the sample by ~15 μm increments 6 times, and ~90 μm increment once, resulting in 8 different relative z positions of approximately 0 μm, 15 μm, 30 μm, 45 μm, 60 μm, 75 μm, 90 μm and 180 μm. We refer to these positions as the 1st, 2nd, …, 8th heights, respectively. The holograms at the 1st, 7th and 8th heights were used to initially calculate the optical phase at the 7th height, using the transport of intensity equation (TIE) through an elliptic equation solver[43], implemented in MATLAB. Combined with the square-root of the hologram intensity acquired at the 7th height, the resulting complex field is used as an initial guess for the subsequent iterations of the multi-height phase recovery. This initial guess is digitally refocused to the 8th height, where the amplitude of the guess is averaged with the square-root of the hologram intensity acquired at the 8th height, and the phase information is kept unchanged. This updating procedure is repeated at the 7th, 6th, …, 1st heights, which defines one iteration of the algorithm. Usually, 10-20 iterations give satisfactory reconstruction results. However, in order to ensure the optimality of the phase retrieval for the training of the network, the algorithm iterated 50 times, after which the complex field is back-propagated to the sample plane, yielding the amplitude and phase, or, real and imaginary images of the sample. These resulting complex-valued images are used to train the network and provide comparison images to the blind testing of the network output.

**Generation of training data**

To generate the training data for the deep neural network, each resulting complex-valued object image from the multi-height phase recovery algorithm as well as the corresponding single hologram back-propagation image (which includes the twin-image and self-interference related spatial artifacts) are divided into 5×5 sub-tiles, with an overlap amount of 400 pixels in each dimension. For each sample type, this results in a dataset of 150 image pairs (i.e., complex-valued input images to the network and the corresponding multi-height reconstruction images), which are divided into 100 image pairs for training, 25 image pairs for validation and 25 image pairs for blind testing. The average computation time for the training of each sample type specific deep neural network (which needs to be done only once) was approximately 14.5 hours, while it increased to approximately 22 hours for the universal deep neural network (refer to the Supplementary Information for additional details).

**Speeding up holographic image reconstruction using GPU programming**

As further detailed in Supplementary Information, the pixel super-resolution and multi-height phase retrieval algorithms are implemented in C/C++ and accelerated using CUDA Application Program Interface (API). These algorithms are run on a laptop computer using a single NVIDIA GTX 1080 graphics card. The basic image operations are implemented using customized kernel functions and are tuned to optimize the GPU memory access based on the access patterns of individual operations. GPU-accelerated libraries such as cuFFT[48] and Thrust[49] are utilized for development productivity and optimized performance. The TIE initial guess is generated using a MATLAB-based implementation, which is interfaced using MATLAB C++ engine API, allowing the overall algorithm to be kept within a single executable after compilation.

**Sample preparation**

*Breast tissue slide*: Formalin-fixed paraffin-embedded (FFPE) breast tissue is sectioned into 2 μm slices and stained using haemotoxylin and eosin (H&E). The de-identified and existing slides are obtained from the Translational Pathology Core Laboratory at UCLA.

*Pap smear:* De-identified and existing Papanicolaou smear slides were obtained from UCLA Department of Pathology.

*Blood smear:* De-identified blood smear slides are purchased from Carolina Biological (Item # 313158).

# Figures and Tables

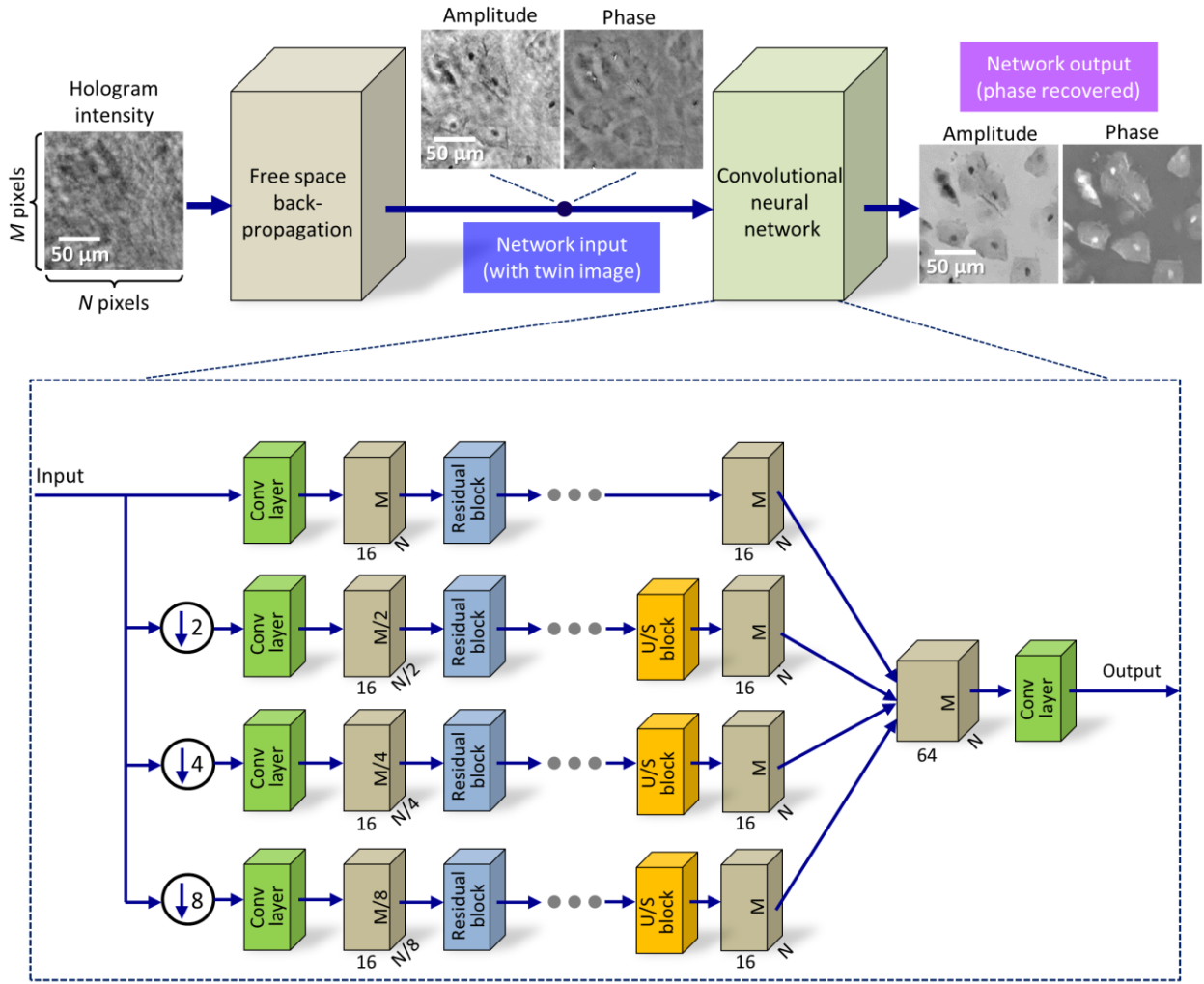

**Fig. 1.** Following its training phase, the deep neural network blindly outputs artifact-free phase and amplitude images of the object using only one hologram intensity. This deep neural network is composed of convolutional layers, residual blocks and upsampling blocks (see the Supplementary Information for additional details) and it rapidly processes a complex-valued input image in a parallel, multi-scale manner.

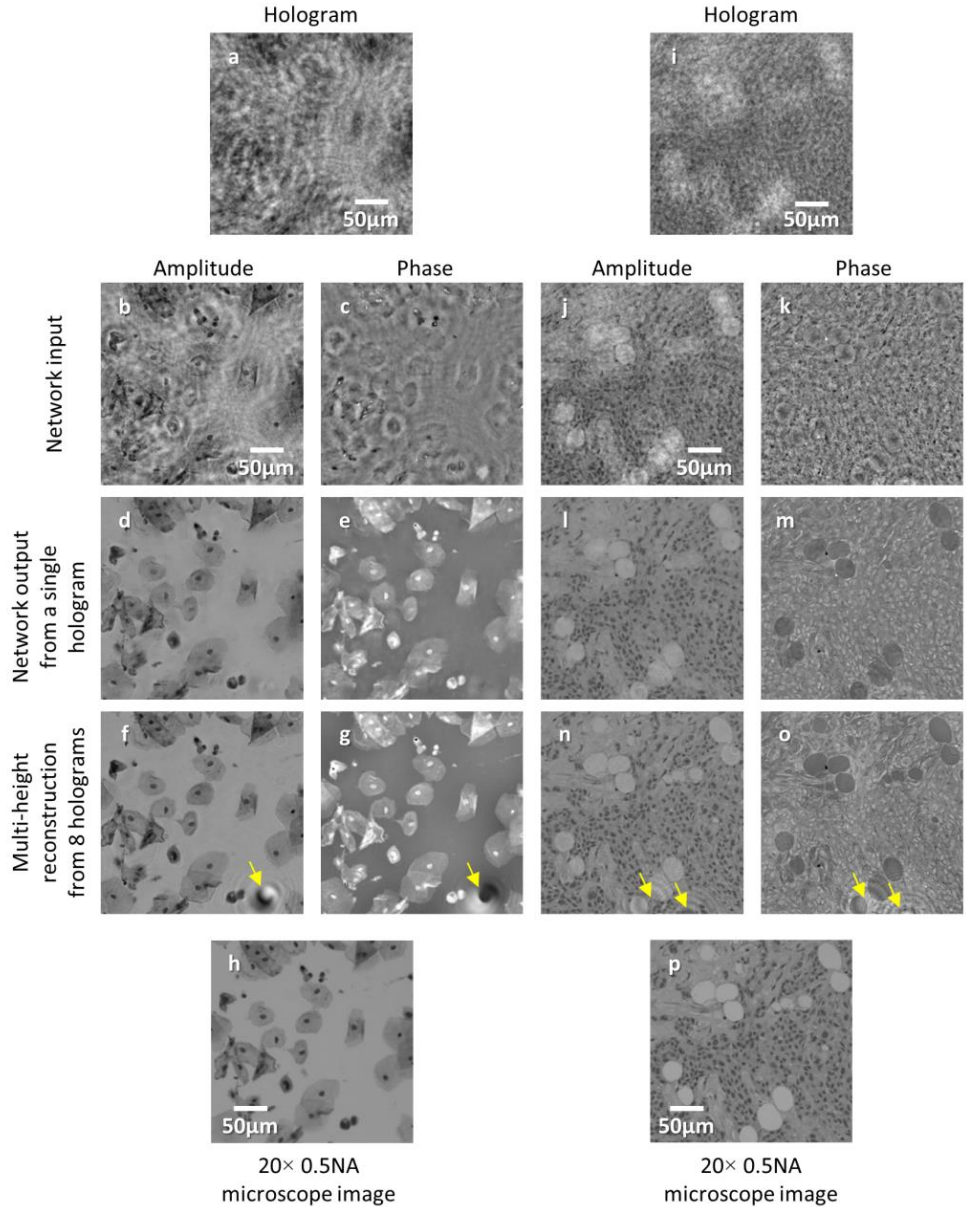

**Fig. 2.** Comparison of the holographic reconstruction results for different types of samples: (**a-h**) Pap smear, (**i-p**) breast tissue section. **a**, **i**, zoomed-in regions of interest from the acquired holograms. **b, c, j, k** amplitude and phase images resulting from free-space backpropagation of a single hologram intensity, shown in **a** and **i**, respectively. These images are contaminated with twin-image and self-interference related spatial artifacts due to the missing phase information at the hologram detection process. **d**, **e**, **l**, **m**, corresponding amplitude and phase images of the same samples obtained by the deep neural network, demonstrating the blind recovery of the complex object image without twin-image and self-interference artifacts using a single hologram. **f**, **g**, **n**, **o**, amplitude and phase images of the same samples reconstructed using multi-height phase retrieval with 8 holograms acquired at different sample-to-sensor distances. **h**, **p**, corresponding bright-field microscopy images of the same samples, shown for comparison. The yellow arrows point to artifacts in **f, g, n, o** (due to out-of-focus dust particles or other unwanted objects) and are significantly suppressed by the network reconstruction as shown in **d**, **e**, **l**, **m**.

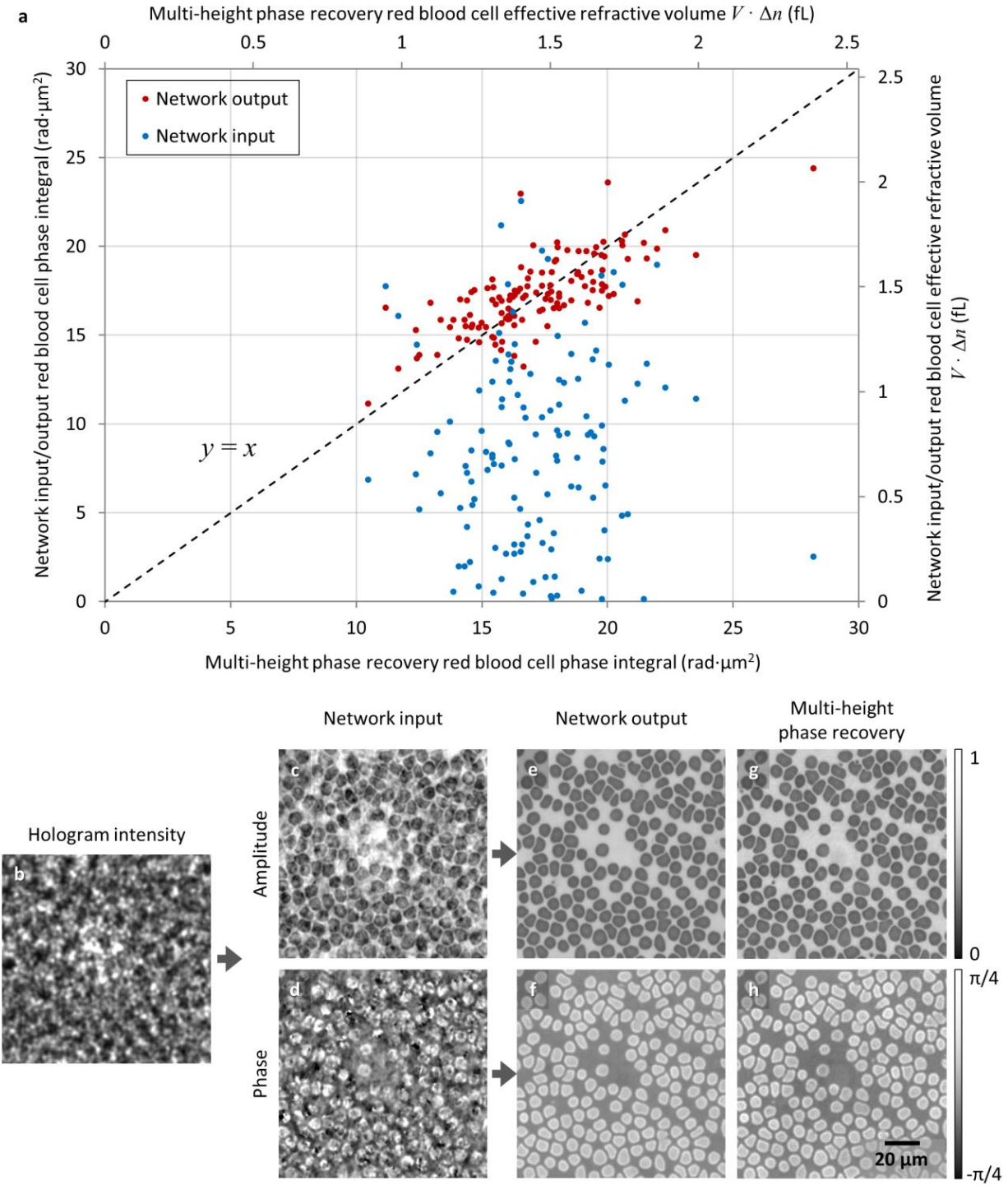

**Fig. 3.** Red blood cell volume estimation using our deep neural network-based phase retrieval. The deep neural network output (**e**, **f**), given the input (**c**, **d**) that is obtained from a single hologram intensity (**b**), shows a good match to the multi-height phase recovery based cell volume estimation results (**a**), calculated using $N_{holo} = 8$ (**g**, **h**). Similar to the yellow arrows shown in Fig. 2(**f**, **g**, **n**, **o**), the multi-height phase recovery results exhibit an out-of-focus fringe artifact at the center of the field-of-view in (**g**, **h**). Refer to the Supplementary Information for the calculation of the effective refractive volume of cells.

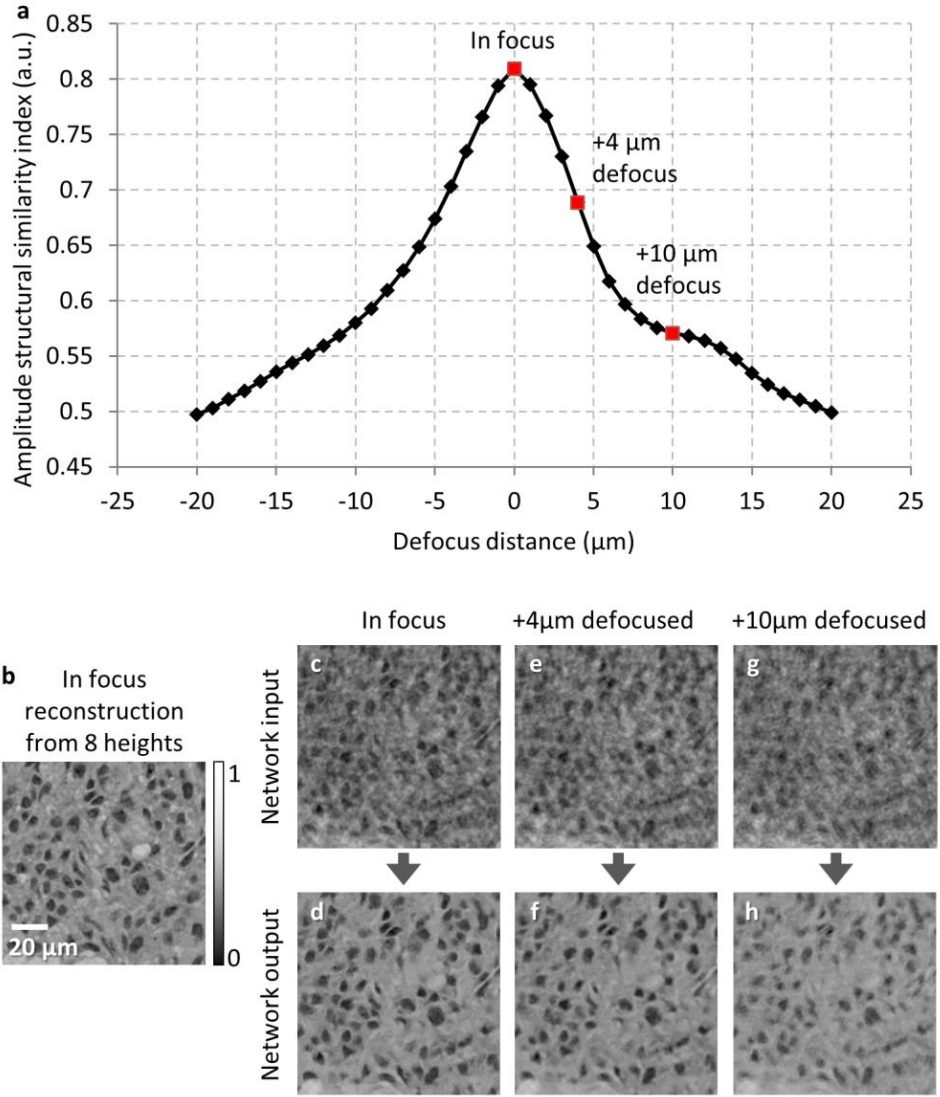

**Fig. 4.** Estimation of the depth defocusing tolerance of the deep neural network. **a**, SSIM index for the neural network output images, when the input image is defocused, i.e., deviates from the optimal focus used in the training of the network. The SSIM index compares the network output images, e.g., **d**, **f**, **h**, with respect to the image obtained by using the multi-height phase recovery algorithm with $N_{holo} = 8$, which is shown in **b**.

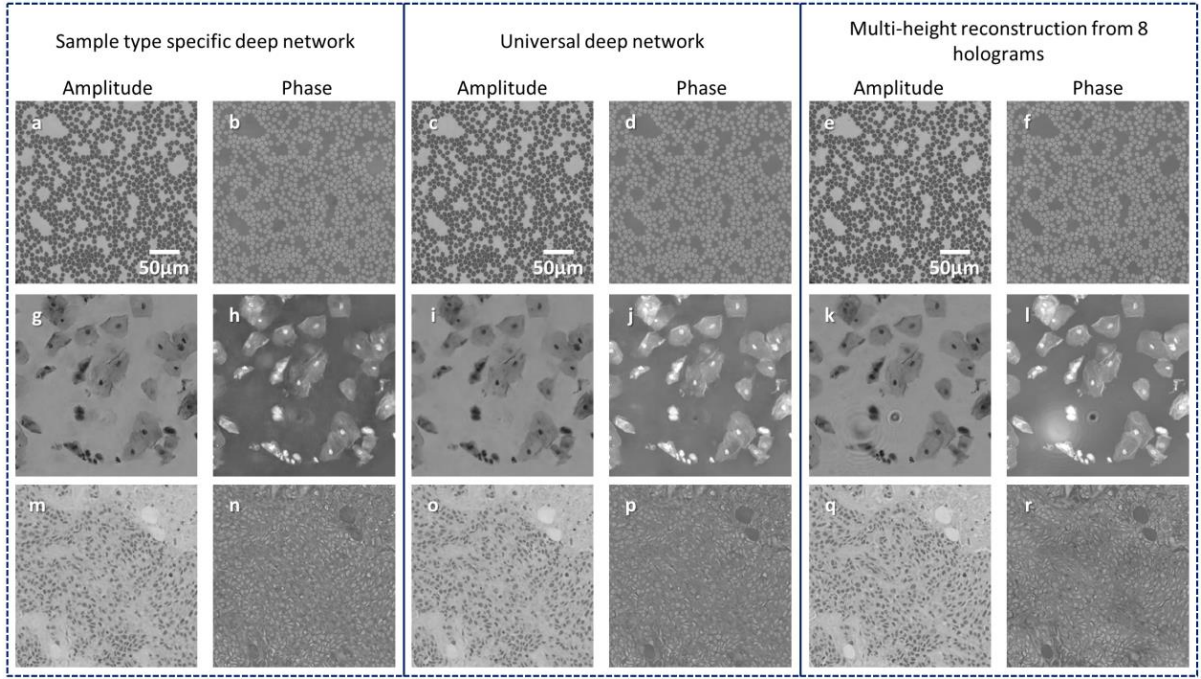

**Fig. 5.** Comparison of holographic image reconstruction results for sample type specific and universal deep networks for different types of samples. Deep neural network results used a single hologram intensity as input, whereas $N_{holo}$ = 8 was used for the column on the right. (**a-f**) Blood smear. (**g-l**) Papanicolaou smear. (**m-r**) Breast tissue section.

| Sample type \ Reconstruction method | Deep network input ($N_{holo}=1$) | Deep network output (Sample type specific) ($N_{holo}=1$) | Deep network output (Universal) ($N_{holo}=1$) | Multi-height phase-recovery ($N_{holo}=2$) | Multi-height phase-recovery ($N_{holo}=3$) | Multi-height phase-recovery ($N_{holo}=4$) | Multi-height phase-recovery ($N_{holo}=5$) | Multi-height phase-recovery ($N_{holo}=6$) | Multi-height phase-recovery ($N_{holo}=7$) | Multi-height phase-recovery ($N_{holo}=8$) |
|---|---|---|---|---|---|---|---|---|---|---|
| Pap smear *real part* | 0.726 | **0.895** | **0.893** | 0.875 | 0.922 | 0.954 | 0.979 | 0.985 | 0.986 | 1 |
| Pap smear *imaginary part* | 0.431 | **0.870** | **0.870** | 0.840 | 0.900 | 0.948 | 0.979 | 0.986 | 0.987 | 1 |
| Blood smear *real part* | 0.701 | **0.942** | **0.951** | 0.890 | 0.942 | 0.962 | 0.970 | 0.975 | 0.977 | 1 |
| Blood smear *imaginary part* | 0.048 | **0.930** | **0.925** | 0.46 | 0.849 | 0.907 | 0.935 | 0.938 | 0.955 | 1 |
| Breast tissue *real part* | 0.826 | **0.916** | **0.921** | 0.931 | 0.955 | 0.975 | 0.981 | 0.983 | 0.984 | 1 |
| Breast tissue *imaginary part* | 0.428 | **0.912** | **0.916** | 0.911 | 0.943 | 0.970 | 0.979 | 0.981 | 0.982 | 1 |

**Table 1.** Comparison of the SSIM index values for the deep neural network output images obtained from a single hologram intensity (for both the sample type specific and universal networks) against multi-height phase retrieval results for different number of input holograms ($N_{holo}$), corresponding to Pap smear samples, breast tissue histopathology slides and blood smear samples. In each case, the SSIM index is separately calculated for the *real* and *imaginary* parts of the resulting complex-valued image with respect to the multi-height phase recovery result for $N_{holo}=8$, and by definition, the last column on the right has an SSIM index of 1. Due to the presence of twin-image and self-interference artifacts, the first column formed by the input images has, by far, the worst performance.

|  | Deep network output (Sample type specific) ($N_{holo}$=1) | Deep network output (Universal) ($N_{holo}$=1) | Multi-height phase-recovery ($N_{holo}$=2) | Multi-height phase-recovery ($N_{holo}$=3) | Multi-height phase-recovery ($N_{holo}$=4) | Multi-height phase-recovery ($N_{holo}$=5) | Multi-height phase-recovery ($N_{holo}$=6) | Multi-height phase-recovery ($N_{holo}$=7) | Multi-height phase-recovery ($N_{holo}$=8) |
|---|---|---|---|---|---|---|---|---|---|
| Runtime (sec) | 6.45 | 7.85 | 23.20 | 28.32 | 32.11 | 35.89 | 38.28 | 43.13 | 47.43 |

**Table 2.** Comparison of the holographic image reconstruction runtime for a field of view of ~1 mm$^2$ for different phase recovery approaches. All the reconstructions were performed on a laptop using a single GPU. Out of the 6.45 sec and 7.85 sec required for image reconstruction from a single hologram intensity using sample type specific and universal neural networks, respectively, the deep neural network processing time is 3.11 sec for the sample type specific network and 4.51 sec for the universal network, while the rest (i.e., 3.34 sec for the preprocessing stages) is used for other operations such as pixel super-resolution, auto-focusing and free space back-propagation.

# Supplementary Information

# Phase recovery and holographic image reconstruction using deep learning in neural networks

Yair Rivenson[1,2,3]†, Yibo Zhang[1,2,3]†, Harun Günaydın[1], Da Teng[1,4], Aydogan Ozcan[1,2,3,5]*

[1]*Electrical Engineering Department, University of California, Los Angeles, CA, 90095, USA.*

[2]*Bioengineering Department, University of California, Los Angeles, CA, 90095, USA.*

[3]*California NanoSystems Institute (CNSI), University of California, Los Angeles, CA, 90095, USA.*

[4]*Computer Science Department, University of California, Los Angeles, CA, 90095, USA*

[5]*Department of Surgery, David Geffen School of Medicine, University of California, Los Angeles, CA, 90095, USA.*

*Correspondence: ozcan@ucla.edu

**Network architecture**

Our deep neural network architecture is detailed in Fig. 1 and Supplementary Figs. 1-2. The real and imaginary parts of the back-propagated hologram intensity are used as two input image channels to the network, each with a size of $M \times N$ pixels (e.g., $M = 1392$, $N = 1392$). These two channels of the network are then used *simultaneously* as input to 4 convolutional layers. The output of each convolutional layer is 16 channels (feature maps), each with a size of $M \times N$ pixels, which was empirically determined to balance the deep network size/compactness and performance. The value of $x,y$-th pixel in the $j$-th feature map in the $i$-th convolutional layer is given by $v_{i,j}^{x,y}$:[1]

$$v_{i,j}^{x,y} = \sum_{r} \sum_{p=0}^{P-1} \sum_{q=0}^{Q-1} w_{i,j,r}^{p,q} v_{i-1,r}^{x+p,y+q} + b_{i,j} \tag{s1}$$

where $b_{i,j}$ is a common bias term for the $j$-th feature map, $r$ indicates the set of the feature maps in the $i$-1 layer (which is 2, for the first convolutional layer), $w_{i,j,r}^{p,q}$ is the value of the convolution kernel at the $p,q$-th position, $P$ and $Q$ define the size of the convolutional kernels, which is $3 \times 3$ throughout the network in our implementation.

For object type-based deep networks, the output of these 4 convolutional layers is then downsampled by $\times 1$, $\times 2$, $\times 4$, $\times 8$, creating 4 different data flow paths, with 16 channels and spatial dimensions of $M \times N$, $M/2 \times N/2$, $M/4 \times N/4$ and $M/8 \times N/8$, respectively. This multi-scale data processing scheme was created to allow the network to learn how to suppress the twin-image and self-interference artifacts, created by objects with different feature sizes. The output of these downsampling operators is followed by 4 residual blocks[2], each composed of 2 convolutional layers and 2 activation functions, which we chose to implement as rectified linear units (ReLU), i.e., $\text{ReLU}(x) = \max(0, x)$. Residual blocks create a shortcut between the block's input and output, which allows a clear path for information flow between layers[3]. This has been demonstrated to speed up the convergence of the training phase of the deep neural network. Following the 4 residual blocks, data at each scale are upsampled to match the original data dimensions. Each upsampling block (i.e., U/S block in Supplementary Fig. 1) contains a convolutional layer that takes 16 channels, each with $M/L \times N/L$ pixels as input, and outputs 64 channels each with $M/L \times N/L$ pixels ($L=2, 4, 8$). This is followed by a ReLU operation and an upsampling layer, which is schematically detailed in Supplementary Fig. 2. This layer learns to upsample a 64 channel input (each with $M/L \times N/L$ pixels) to a 16 channel output (each with $2M/L \times 2N/L$ pixels). This upsampling process is being performed once, twice, or three times, for the $\times 2$, $\times 4$, $\times 8$ spatially downsampled network inputs, respectively (see Supplementary Fig. 1). The output of each one of these 4 different dataflow paths (with 16 channels, $M \times N$ pixels, following the upsampling stage) is concatenated to a 64 channels input, which results in 2 channels: one for the real part and one for the imaginary part of the object image, each having $M \times N$ pixels. For the universal deep network, we kept the same architecture; however, we increased the number of channels in the output of each convolutional layer by two-fold, i.e., from 16 to 32 in the residual blocks.

To train the network, we minimized the average of the mean-squared-errors of the real and imaginary parts of the network output with respect to the real and imaginary parts of the object's ground truth images, obtained using multi-height phase retrieval with 8 holograms recorded at different sample-to-sensor distances (also see the Methods section of the main text). This loss function over a mini-batch of $K$ input patches (images) is calculated as:

$$Loss(\Theta) = \frac{1}{2K}\sum_{k=1}^{K}\left\{\frac{1}{M\times N}\sum_{m=1}^{M\times N}\sum_{n=1}^{M\times N}\left\|Y_{\text{Re},m,n,k}^{\Theta} - Y_{\text{Re},m,n,k}^{GT}\right\|^2 + \frac{1}{M\times N}\sum_{m=1}^{M\times N}\sum_{n=1}^{M\times N}\left\|Y_{\text{Im},m,n,k}^{\Theta} - Y_{\text{Im},m,n,k}^{GT}\right\|^2\right\} \quad \text{(s2)}$$

where $k$ is the $k$-th image patch, $Y_{\text{Re},m,n,k}^{\Theta}$, $Y_{\text{Im},m,n,k}^{\Theta}$ denote the $m,n$-th pixel of real and imaginary network outputs, respectively, and $Y_{\text{Re},m,n,k}^{GT}$, $Y_{\text{Im},m,n,k}^{GT}$ denote the $m,n$-th pixel of real and imaginary parts of the training (i.e., ground truth) labels, respectively. The network's parameter space (e.g., kernels, biases, weights) is defined by $\Theta$ and its output is given by $[Y_{\text{Re}}^{\Theta}, Y_{\text{Im}}^{\Theta}] = F(X_{\text{Re},input}, X_{\text{Im},input}; \Theta)$, where $F$ defines the deep neural network's operator on the back propagated complex field generated from a single hologram intensity, divided into real and imaginary channels, $X_{\text{Re},input}, X_{\text{Im},input}$, respectively. Following the estimation of the loss function, the resulting error in the network output is back-propagated through the network and the Adaptive Moment Estimation[4] (ADAM) based optimization is used to tune the network's parameter space, $\Theta$, with a learning rate of $10^{-4}$. For the sample type specific network training, we used a batch size of $K=2$ and an image size of 1392×1392 pixels. For the universal deep network, we divided the image dataset to 256×256-pixel patches (with an overlap of 20% between the patches) and a mini-batch size of $K=30$ (see Supplementary Fig. 3). All the convolutional kernel entries are initialized using a truncated normal distribution. All the network bias terms, $b_{i,j}$, are initialized to 0. In case the size of the input image is not divisible by 8, zero padding is performed on it such that it becomes divisible by 8.

**Network implementation details**
For our programming, we used Python version 3.5.2, and the deep neural network was implemented using TensorFlow framework version 1.1.0 (Google). We used a laptop computer with Core i7-6700K CPU @ 4GHz (Intel) and 64GB of RAM, running a Windows 10 operating system (Microsoft). The network training was performed using GeForce GTX 1080 (NVidia) Dual Graphical Processing Units (GPUs). The testing of the network was performed on *a single GPU* to provide a fair comparison against multi-height phase retrieval CUDA implementation, as summarized in Table 2 (main text).

**Optical set-up**
Our experimental set-up (Supplementary Fig. 4) includes a laser source (SC400, Fianium Ltd., Southampton, UK) filtered by an acousto-optic tunable filter and coupled to a single mode optical fiber to provide partially coherent illumination with a spectral bandwidth of ~2.5 nm. A CMOS image sensor with 1.12 μm pixel size and 16.4 Megapixel (IMX081, Sony Corp., Japan) is used to capture the holographic images. The distance from the optical fiber tip to the sample is between 7 and 15 cm, such that the light that is incident on the sample can be considered a quasi-plane wave. The distance from the sample to the image sensor plane is approximately 300-700 μm. This unit magnification geometry results in a large field of view that is equal to the image sensor's active area. The image sensor was mounted on a 3D positioning stage (NanoMax 606, Thorlabs Inc., New Jersey, US), which moved it in *x* and *y* directions in sub-pixel-size steps to implement pixel super-resolution (PSR). The image sensor was also shifted in the z direction with step sizes of a few tens of microns to perform multi-height phase recovery to generate training data for the neural network. A custom-written LabVIEW program implemented on a desktop computer was used to control and automate all of these components as part of the imaging set-up.

**Pixel super resolution (PSR)**

In order to mitigate the spatial undersampling caused by the relatively large pixel pitch of the image sensor chip (~1.12 µm), multiple subpixel-shifted holograms were used to synthesize a higher resolution (i.e., pixel super-resolved) hologram. For this, the image sensor was mechanically shifted by a 6-by-6 rectangular grid pattern in the x-y plane, with increments of 0.37 µm, corresponding to approximately 1/3 of the image sensor's pixel size. A 6-by-6 grid ensured that one color channel of the Bayer pattern could cover its entire period. In an alternative design with a monochrome image sensor (instead of an RGB sensor), only a 3-by-3 grid would be needed to achieve the same PSR factor. For this PSR computation, an efficient non-iterative fusion algorithm was applied to combine these sub-pixel shifted images into one higher-resolution hologram, which preserves the optimality of the solution in the maximum likelihood sense[5]. The selection of which color channel (R, G or B) of the Bayer pattern to use for holographic imaging is based on pixel sensitivity to the illumination wavelength that is used. For example, at ~530 nm illumination, the two green channels of the Bayer pattern were used, and at ~630 nm, the red channel was used.

**Calculation of red blood cell (RBC) phase integral and effective refractive volume**

The relative optical phase delay due to a cell, with respect to the background, can be approximated as:

$$\varphi(x,y) = \frac{2\pi d(x,y) \cdot \Delta n(x,y)}{\lambda} \tag{s3}$$

where $d(x,y)$ is the thickness of the sample (e.g., an RBC) as a function of the lateral position, $\Delta n(x,y) = n(x,y) - n_0$ is the refractive index difference between the sample ($n(x,y)$) and the background medium ($n_0$), $\lambda$ is the illumination wavelength in air. Based on these, we define the phase integral for a given RBC image as:

$$p_i = \left| \int_{S_i} \varphi(x,y) \, ds \right| = \left| \int_{S_i} \frac{2\pi d(x,y) \Delta n(x,y)}{\lambda} \, ds \right| \tag{s4}$$

which calculates the relative phase with respect to the background that is integrated over the area of each RBC (defined by $S_i$), which results in a unit of rad·µm². Let $\Delta n$ represent the average refractive index difference within each cell (with respect to $n_0$), we can then write:

$$p_i = \frac{2\pi \cdot |\Delta n|}{\lambda} \int_{S_i} d(x,y) \cdot ds = \frac{2\pi \cdot |\Delta n|}{\lambda} \cdot V_i \tag{s5}$$

where $V_i$ represents the volume of the *i*th cell. Because the average refractive index of a *fixed and stained* RBC (as one would have in a blood smear sample) is hard to determine or estimate, we instead define *effective refractive volume of an RBC* as:

$$\tilde{V}_i = |\Delta n| \cdot V_i = \frac{p_i \lambda}{2\pi} \tag{s6}$$

which also has the unit of volume (e.g., femtoliter, fL).

**Structural similarity (SSIM) index calculation**

The structural similarity index between two images $I_1$ and $I_2$ can be calculated as[6]:

$$\text{SSIM}(I_1, I_2) = \frac{(2\mu_1\mu_2 + c_1)(2\sigma_{1,2} + c_2)}{(\mu_1^2 + \mu_2^2 + c_1)(\sigma_1^2 + \sigma_2^2 + c_2)} \tag{s7}$$

where $\mu_1$ is the average of $I_1$, $\mu_2$ is the average of $I_2$, $\sigma_1^2$ is the variance of $I_1$, $\sigma_2^2$ is the variance of $I_2$, $\sigma_{1,2}$ is the cross-covariance of $I_1$, and $I_2$. The stabilization constants ($c_1$, $c_2$) prevent division by a small denominator and can be selected as $c_1 = (K_1 L)^2$ and $c_2 = (K_2 L)^2$, where $L$ is the dynamic range of the image and $K_1, K_2$ are both much smaller than 1. SSIM index between two images ranges between 0 and 1 (the latter for identical images).

**Evaluation of scattering strength of the samples**

To evaluate the validity of the weakly scattering condition, i.e., $|a(x,y)| \ll A$ for the samples that we imaged (see the Discussion section of the main text), we took a region of interest for each of the samples that is reconstructed using the multi-height phase recovery, based on 8 hologram heights. After the phase recovery step, we have:

$$u = A + a(x, y) \tag{s8}$$

where $A$ can be estimated by calculating the average value of a background region where no sample is present. After $A$ is estimated, we calculate a normalized complex image $\tilde{u}$,

$$\tilde{u} = \frac{u}{A} = 1 + \frac{a(x, y)}{A} \tag{s9}$$

Next, we define $R$ as the ratio between the root-mean-squared (RMS, or quadratic mean) modulus of the scattered wave $|a(x,y)|$ divided by the reference wave modulus $|A|$, to obtain:

$$R = \frac{\left\langle |a(x, y)|^2 \right\rangle^{1/2}}{|A|} = \left\langle |\tilde{u} - 1|^2 \right\rangle^{1/2} \tag{s10}$$

where $\langle \cdot \rangle$ denotes 2D spatial averaging operation. This ratio, $R$, is used to evaluate the validity of the weakly scattering condition for our samples, and is found to be 0.28, 0.34, and 0.37 for the breast tissue, Pap smear and blood smear samples that we imaged, respectively (see the Discussion section).

**Calculation of the sample-to-sensor distance**

The relative separation between successive image sensor heights (or hologram planes) needs to be estimated in order to successfully apply the TIE and multi-height phase recovery algorithms, and the absolute $z_2$ distance (i.e., the sample-to-sensor distance, see Supplementary Fig. 4) is needed for the final back-propagation of the recovered complex wave onto the sample plane. Estimating the relative z-separation is done by using an autofocusing algorithm based on the axial magnitude differential[7]. For computational efficiency, first a coarse scan is done between 100 μm and 800 μm with a step size of 10 μm. Then, around the minimum that is found by this coarse scan, a golden section search algorithm[8] is applied to locate the minimum with a final precision of 0.01 μm. The absolute $z_2$ is refined after the convergence of the multi-height phase recovery algorithm by refocusing the phase-recovered hologram near the previously found focus point.

**Supplementary Figures**

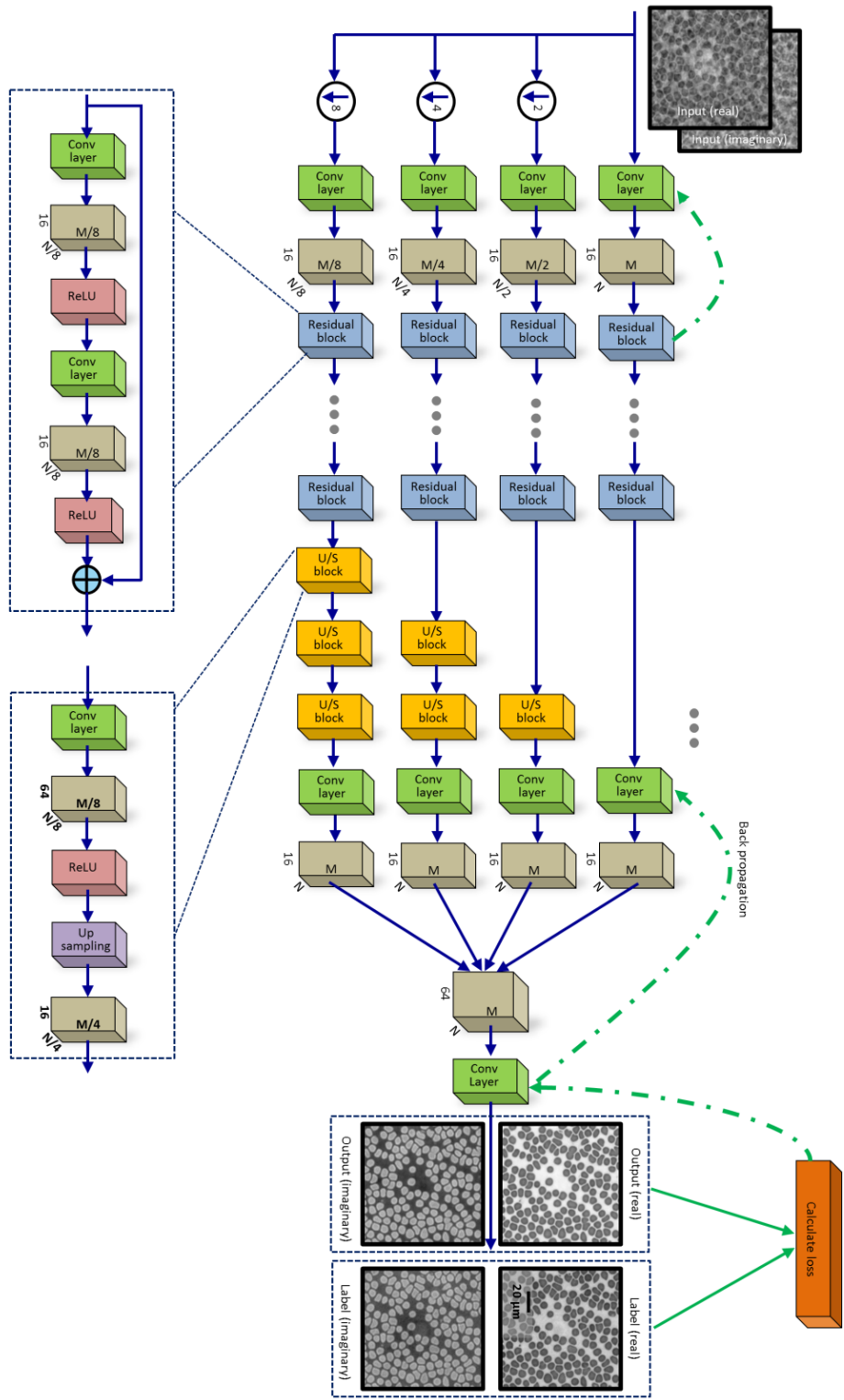

**Supplementary Fig. 1.** Architecture of our deep neural network and its training. The neural network is composed of convolutional layers (i.e., conv layers), upsampling blocks (U/S blocks) and nonlinear activation functions (ReLU). Also see Supplementary Fig. 2.

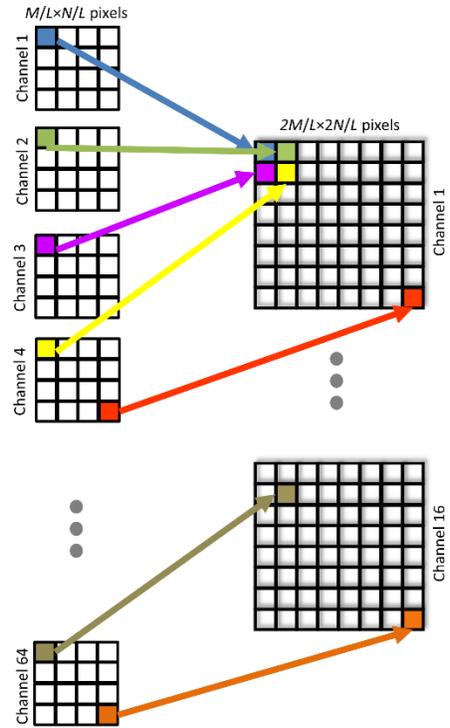

**Supplementary Fig. 2.** Detailed schematics of the upsampling layer of our deep neural network.

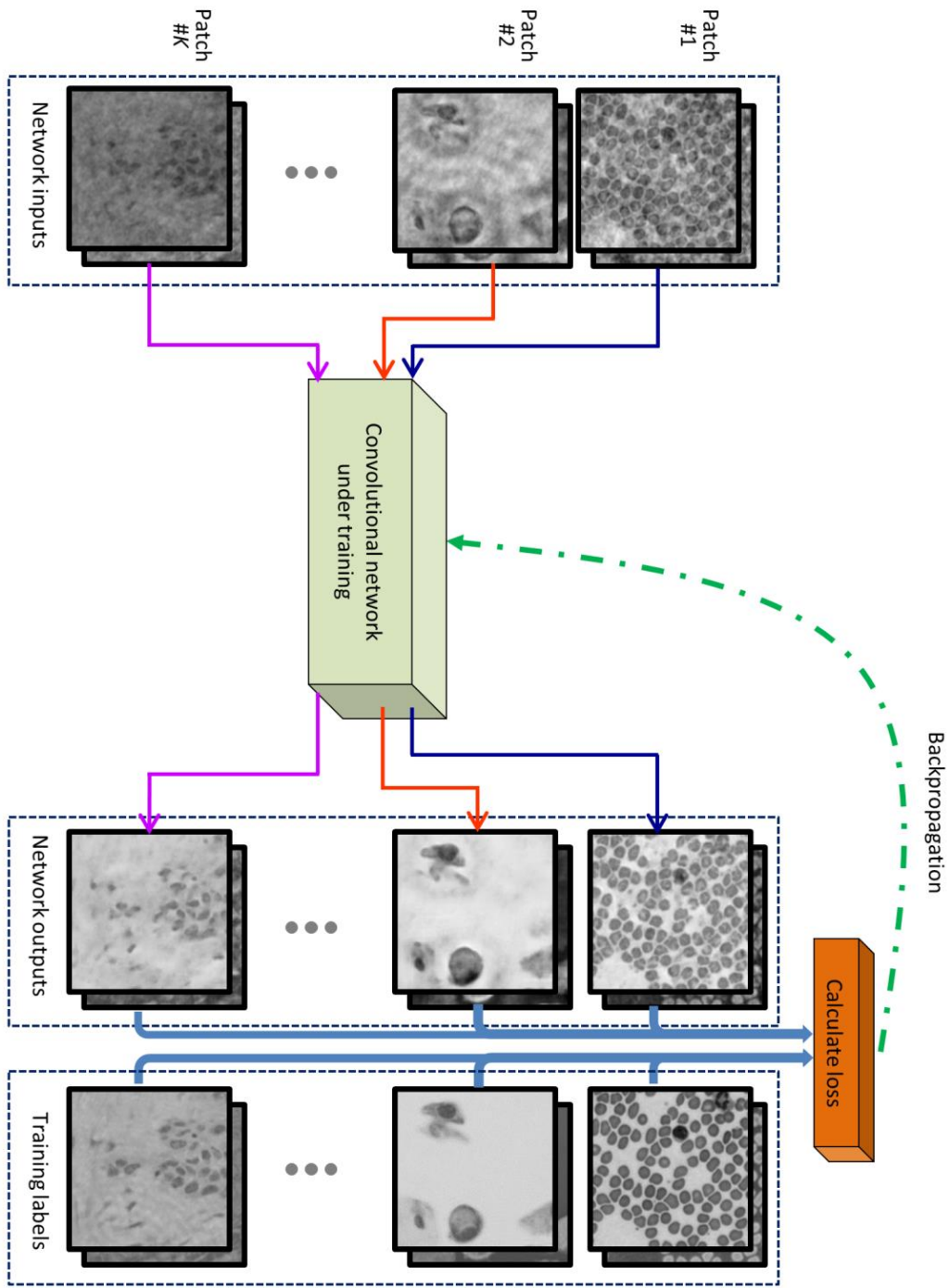

**Supplementary Fig. 3.** Training of the universal deep neural network that can reconstruct different types of objects.

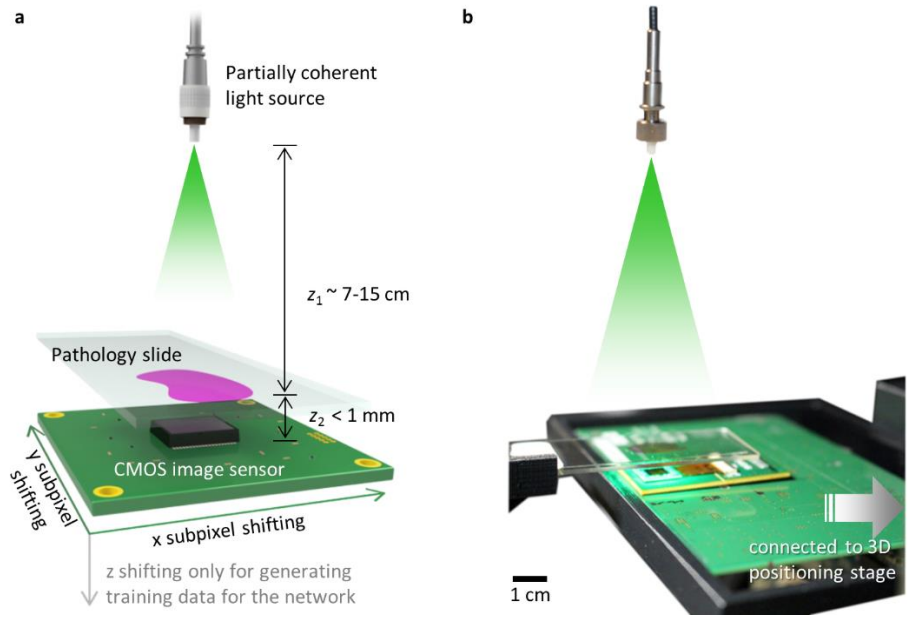

**Supplementary Fig. 4.** Holographic imaging setup. **a,** schematics of the optical set-up. **b,** a photograph of the same setup.